\documentclass[lettersize,journal]{IEEEtran}
\usepackage{amsmath,amsfonts}
\usepackage{algorithmic}
\usepackage{algorithm}
\usepackage{array}
\usepackage[caption=false,font=normalsize,labelfont=sf,textfont=sf]{subfig}
\usepackage{textcomp}
\usepackage{stfloats}
\usepackage{url}
\usepackage{verbatim}
\usepackage{graphicx}
\usepackage{booktabs}
\usepackage{tabularx}
\usepackage[table]{xcolor} 
\usepackage{siunitx}      
\usepackage{tabularx}
\usepackage{cite}
\usepackage{booktabs} 
\usepackage{multirow}
\usepackage{caption}
\usepackage{xcolor}
\usepackage[colorlinks=true, citecolor=blue]{hyperref}

\begin{document}

\title{Rad-VLSM:  A Cross-Modal Framework with Semantics-Assisted Prompting for Medical Segmentation and Diagnosis}

\author{Fengyi Zhang\textsuperscript{*}, \IEEEmembership{Student Member, IEEE}, 
        Xujie Zeng\textsuperscript{*}, \IEEEmembership{Student Member, IEEE}, Mohan Liu\textsuperscript{*}, Zengyi Wang, \IEEEmembership{Student Member, IEEE} and Yalong Jiang, \IEEEmembership{Member, IEEE}}

\maketitle

\begin{abstract}
Medical image segmentation is more clinically valuable when it supports diagnosis rather than merely producing lesion masks. However, diagnostically relevant lesion cues are often subtle and localized, while existing models may be distracted by background tissues, acoustic artifacts, and irrelevant visual correlations. To address this problem, we propose Rad-VLSM, a two-stage cross-modal framework for semantics-assisted lesion focusing, robust segmentation, and visually grounded diagnosis. In the first stage, a BLIP-2-based vision-language alignment module identifies lesion-related candidate regions under semantic guidance and converts them into box prompts. In the second stage, these prompts are fed into a SAM-based multitask network, where a multi-candidate region aggregation strategy improves prompt stability and guides lesion segmentation. The predicted masks are then used as spatial priors for diagnosis, and a visual-radiomics fusion head integrates lesion-aware visual features with selected radiomics descriptors. By using semantic information for localization rather than direct prediction, Rad-VLSM reduces text-to-diagnosis dependence and grounds diagnosis in lesion-level evidence. Experiments on a private clinical breast ultrasound dataset and public benchmarks show that Rad-VLSM achieves strong segmentation and diagnostic performance with favorable generalization.
\end{abstract}


\section{Introduction}
\begin{figure}[htbp]
    \centering
      \includegraphics[width=\linewidth]{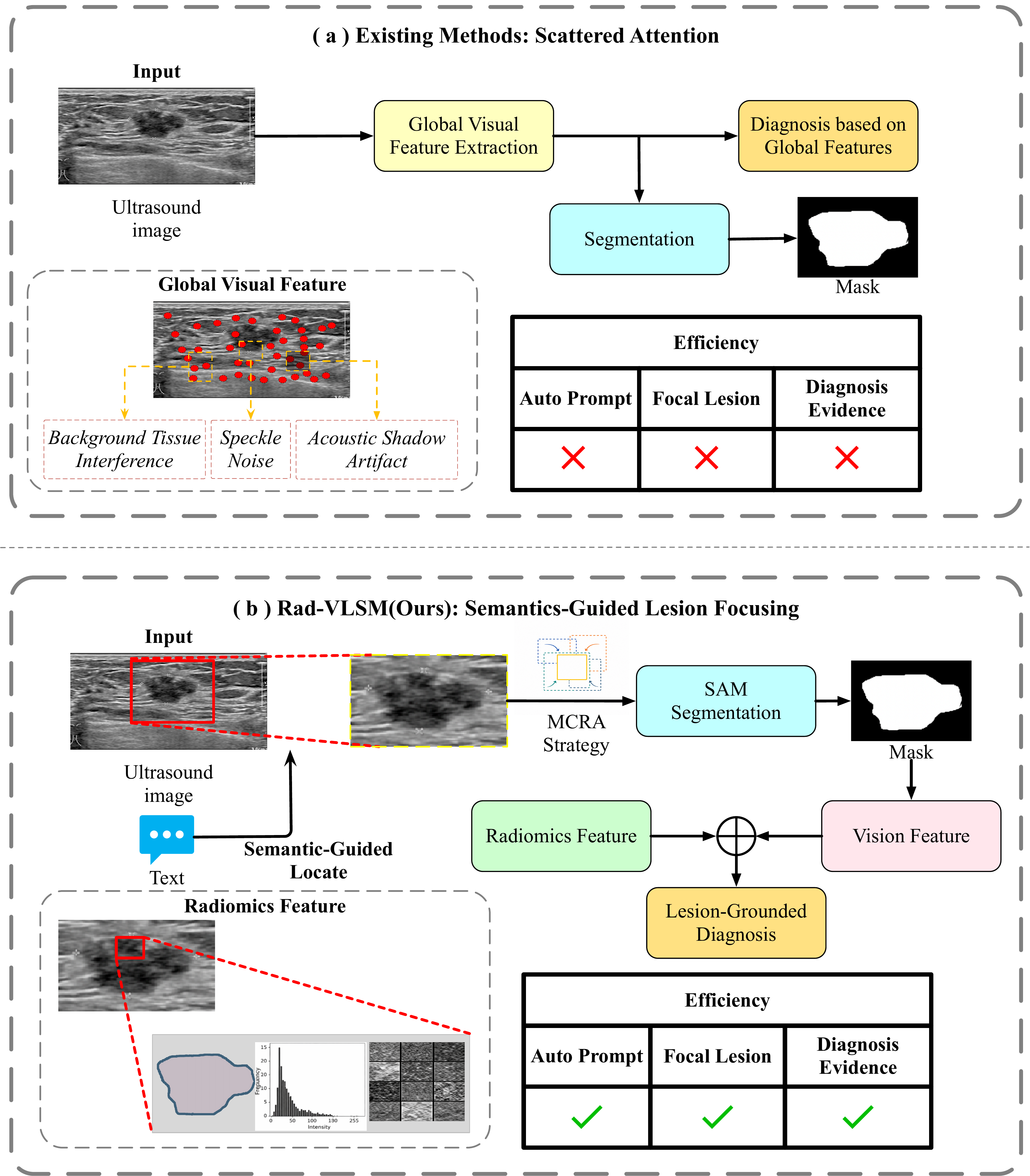}
    \caption{ Comparison between existing methods and ours Rad-VLSM.
    (a) Existing segmentation--classification methods often rely on whole-image representations, making diagnosis vulnerable to Interfering information.
    (b) Rad-VLSM progressively focuses on lesion regions and performs lesion-grounded diagnosis using mask-guided visual and radiomics evidence.}
    \label{fig:overview}
\end{figure}
Medical imaging plays a pivotal role in disease screening, subtype diagnosis, and treatment planning, while accurate lesion segmentation remains one of its most fundamental technical components \cite{ronneberger2015u},\cite{cciccek20163d},\cite{isensee2021nnunet}. 

Recent deep learning models have achieved high-quality masks across a variety of medical segmentation tasks, and the emergence of foundation models has further improved their adaptability across imaging scenarios \cite{chen2024transunet},\cite{kirillov2023segment},\cite{ma2024segment}. However, in real clinical workflows, a high-quality mask is not the ultimate goal. What clinicians truly need is a model that can not only delineate the lesion precisely, but also make a reliable diagnostic judgment and explicitly indicate the evidence supporting that judgment, thereby forming a diagnostic loop grounded in spatial evidence \cite{chen2023medrpg},\cite{zou2025uncertainty},\cite{borys2023xai}. Therefore, beyond developing isolated segmentation or classification models, it is clinically meaningful to build a unified framework that connects lesion localization, segmentation, and diagnosis\cite{yahyatabar2023joint},\cite{von2021multitask},\cite{zhao2023multi}.

This challenge becomes even more pronounced in ultrasound-based lesion analysis. In breast ultrasound images,  lesions often occupy only a small fraction of the image and are frequently accompanied by weak boundaries, low contrast, heterogeneous internal echoes, speckle noise, and acoustic shadows, making reliable lesion-level representation difficult \cite{ilesanmi2021methods},\cite{wang2024frequency},\cite{huang2022trustworthy}. Meanwhile, surrounding glandular tissues and extra-lesion contexts may introduce confounding visual patterns. As a result, models relying primarily on global image-level representations may be distracted by irrelevant regions, leading to the dilution of lesion-specific discriminative cues\cite{byra2022busxai,nastase2024role},as shown in Fig.~\ref{fig:overview}(a) .

Promptable segmentation models such as SAM and MedSAM provide a promising way to extract lesion regions through spatial prompts \cite{kirillov2023segment,mazurowski2023segment,ma2024segment}. 
Recent studies have further explored medical SAM adaptation from multiple perspectives, including uncertainty-guided multi-prompt adaptation, SAM-based semi-supervised decoding, lightweight modality-decoupled adaptation, and gaze-assisted prompting \cite{11201277,11217143,11203003,ge2025adaptation}. 
However, their effectiveness still depends heavily on whether the provided prompts can reliably indicate the lesion area. Manual prompts are costly, while automatically generated prompts may be unstable when lesion boundaries are weak or the background is complex \cite{xie2025masksam,huang2024robust}.

Vision-language alignment provides a potential way to transform lesion-related semantic understanding into spatial localization cues \cite{wang2024sam,li2025vision,lu2025integrating}. By associating clinically meaningful lesion descriptions with corresponding local visual patterns, medical vision-language models can help identify lesion-relevant regions and suppress confounding background responses \cite{huang2021gloria,wu2023medklip,chen2023medrpg}, thereby providing a feasible basis for automatic box prompt generation. However, if semantic cues are directly injected into the diagnostic classifier, the model may exploit label-correlated textual semantics rather than grounding its prediction in lesion morphology, internal texture, and peri-lesional context \cite{ong2024shortcut,nguyen2025localizing}. Such semantic shortcut learning may weaken the lesion-centered nature of diagnostic evidence.

Even after lesion regions are localized and segmented, unified diagnosis remains challenged by the ambiguity of fine-grained lesion evidence\cite{wang2024frequency}. In many clinical scenarios, diagnostically important differences are not reflected in obvious global appearance, but in subtle and localized variations in lesion morphology, internal texture, and surrounding context \cite{kundel1990visual},\cite{mayerhoefer2020introduction},\cite{litvin2021radiomics},\cite{nastase2024role}. These cues are often weak, heterogeneous, and difficult to characterize reliably, even when coarse lesion localization is available \cite{mayerhoefer2020introduction},\cite{litvin2021radiomics},\cite{lambin2012radiomics}. As a result, coarse visual cues or weak semantic priors alone are usually insufficient for reliable diagnosis \cite{nastase2024role},\cite{guo2025radiomics},\cite{ye2025multimodalus}.

To address these challenges, we propose a two-stage cross-modal framework that uses semantic priors for prompt generation and visually grounded lesion representations for downstream diagnosis, as shown in Fig.~\ref{fig:overview}(b). In the first stage, a BLIP-2-based vision-language module is adapted to associate  visual patterns with clinically meaningful lesion descriptions, allowing the model to identify lesion-related candidate regions under semantic guidance, reducing the need for manual prompt annotation at inference time. In the second stage, these prompts guide a SAM-based multitask branch to perform lesion segmentation and diagnosis, while avoiding direct fusion of raw text embeddings into the final diagnosis head. To further improve robustness, we introduce a multi-candidate region aggregation (MCRA) strategy to stabilize lesion evidence under prompt variation, and a cross-modal evidence fusion head that combines deep visual features, selected radiomics descriptors, and region-level contextual information for diagnosis. In this way, our framework couples segmentation and diagnosis through lesion-grounded evidence, while reducing the need for manual prompt annotation at inference time, mitigating the impact of imperfect prompt generation, and improving fine-grained lesion characterization.

The main contributions of this work are summarized as follows:
\begin{itemize}
    \item  We propose Rad-VLSM, a semantics-assisted vision-language segmentation and diagnosis framework for medical image analysis. It leverages BLIP-2-based vision-language alignment to associate lesion-related visual patterns with clinically meaningful semantic descriptions, thereby enabling automatic lesion localization and box prompt generation without manual prompt annotation at inference time. 
    
    \item We introduce a MCRA strategy to improve the stability of lesion evidence under prompt variation and coarse localization errors. This strategy produces more reliable lesion-level representations and enhances the robustness of unified diagnosis.
    
    \item We present a lesion-mask-guided diagnosis scheme with visual-radiomics fusion. By integrating implicit visual representations with explicit radiomic features, this scheme improves diagnostic reliability and interpretability.

    \item We rigorously evaluate our framework on a private clinical breast ultrasound dataset provided by Peking Union Medical College Hospital, alongside several public benchmarks. Extensive experiments demonstrate stable and competitive performance in both segmentation and diagnosis, with favorable generalization in challenging breast lesion scenarios.
    
\end{itemize}

\section{related work}

\subsection{Joint Segmentation and Diagnosis}

Existing methods for joint segmentation and diagnosis can be broadly divided into two paradigms~\cite{zhao2023multi}. The first paradigm follows multi-stage pipelines, where lesion regions are first localized or segmented and then used for downstream diagnosis~\cite{bruno2025dual}, \cite{wang2019joint}. For example, Li \emph{et al.}~\cite{li2024joint} proposed a clinical knowledge-aware framework for joint lesion detection and classification in breast ultrasound videos, where lesion localization is explicitly performed to support downstream diagnostic prediction. However, its performance is often highly sensitive to the quality of intermediate localization or segmentation results, making downstream diagnosis prone to error propagation.

The second paradigm adopts shared-backbone multi-task learning, where segmentation and classification are jointly optimized within a unified network~\cite{zhou2021multi},\cite{von2021multitask},\cite{almasni2024unified},\cite{lu2025mtloca},\cite{aumente2025multi}. For instance, Dai \emph{et al.}~\cite{dai2025multi} proposed a multi-task learning network for medical image analysis guided by lesion regions and spatial relationships of tissues, showing that shared feature learning can enhance task collaboration by jointly modeling lesion-aware representations within a unified framework. Nevertheless, the coupling between segmentation and classification in this paradigm is often weak, resulting in limited task interaction and relatively low interpretability.

In contrast, our framework formulates joint segmentation and diagnosis as an evidence-driven pipeline. This design provides a more explicit connection between lesion segmentation and diagnosis, thereby offering a complementary alternative to existing multi-stage and shared-backbone multi-task methods.

\subsection{Medical Vision-Language Models}

Recent advances in medical vision-language models have introduced a promising multimodal paradigm for medical image analysis \cite{lu2025integrating},\cite{li2025vision},\cite{ryu2025vision},\cite{hartsock2024vlmreview}. By aligning visual representations with textual semantics, these models have demonstrated considerable potential in a wide range of applications, including report generation, cross-modal retrieval, visual question answering, and fine-grained lesion understanding \cite{hartsock2024vlmreview},\cite{reale2024vision},\cite{nam2025multimodal}. In the biomedical domain, representative foundation models such as BiomedCLIP\cite{zhang2023biomedclip} have further demonstrated the value of large-scale image-text pretraining for downstream medical tasks. 

Beyond global semantic alignment and report-level understanding, recent studies have increasingly moved toward fine-grained grounding and text-guided localization \cite{chen2024adamatch},\cite{huang2021gloria}. For example, GLoRIA\cite{huang2021gloria} learns global-local medical representations by aligning report words with image subregions. Such progress suggests that language can provide informative semantic priors to complement conventional visual-only learning and enrich lesion understanding. However, the integration of semantic information also brings a critical challenge: once language-derived priors are allowed to directly participate in diagnostic prediction, the model may rely on textual correlations or high-level semantic cues rather than truly grounded lesion-level visual evidence\cite{ong2024shortcut},\cite{lu2025integrating},\cite{li2025vision}. Unlike prior multimodal approaches that directly inject semantic information into the final decision process, our method uses semantic priors for localization and prompt refinement, while keeping final diagnosis centered on lesion-level visual evidence rather than direct text-conditioned prediction.

\subsection{Evidence-Grounded Diagnosis}

Clinically useful diagnosis requires evidence that is explicit, localized, and interpretable~\cite{11142861}. Existing medical image diagnosis models often rely on post hoc visualization tools such as CAMs or saliency maps, which are widely used in medical imaging explainability research \cite{borys2023xai},\cite{muhammad2024unveiling}. However, saliency-based explanations have been shown to be potentially unreliable, as they may fail sanity checks, be sensitive to small perturbations, or insufficiently reflect the model's true decision process\cite{arun2021assessing},\cite{zhang2023revisiting}. 

Radiomics provides a complementary route toward evidence-grounded diagnosis by quantifying lesion phenotype through explicit descriptors of shape, intensity, and texture \cite{lambin2012radiomics},\cite{linton2025radiomics},\cite{ray2025review}. Recent studies have therefore increasingly explored the fusion of deep representations with radiomic features to improve both performance and interpretability, especially in breast ultrasound diagnosis \cite{guo2025radiomics},\cite{ye2025multimodalus},\cite{yang2024dlradiomics}. Building on this idea, our framework integrates lesion-level visual features, candidate-region context, and selected radiomics descriptors into a unified evidence fusion scheme, enabling diagnosis to be supported by richer and more structured lesion evidence.

\section{Method}
\begin{figure*}[t]
    \centering
    \includegraphics[width=\linewidth]{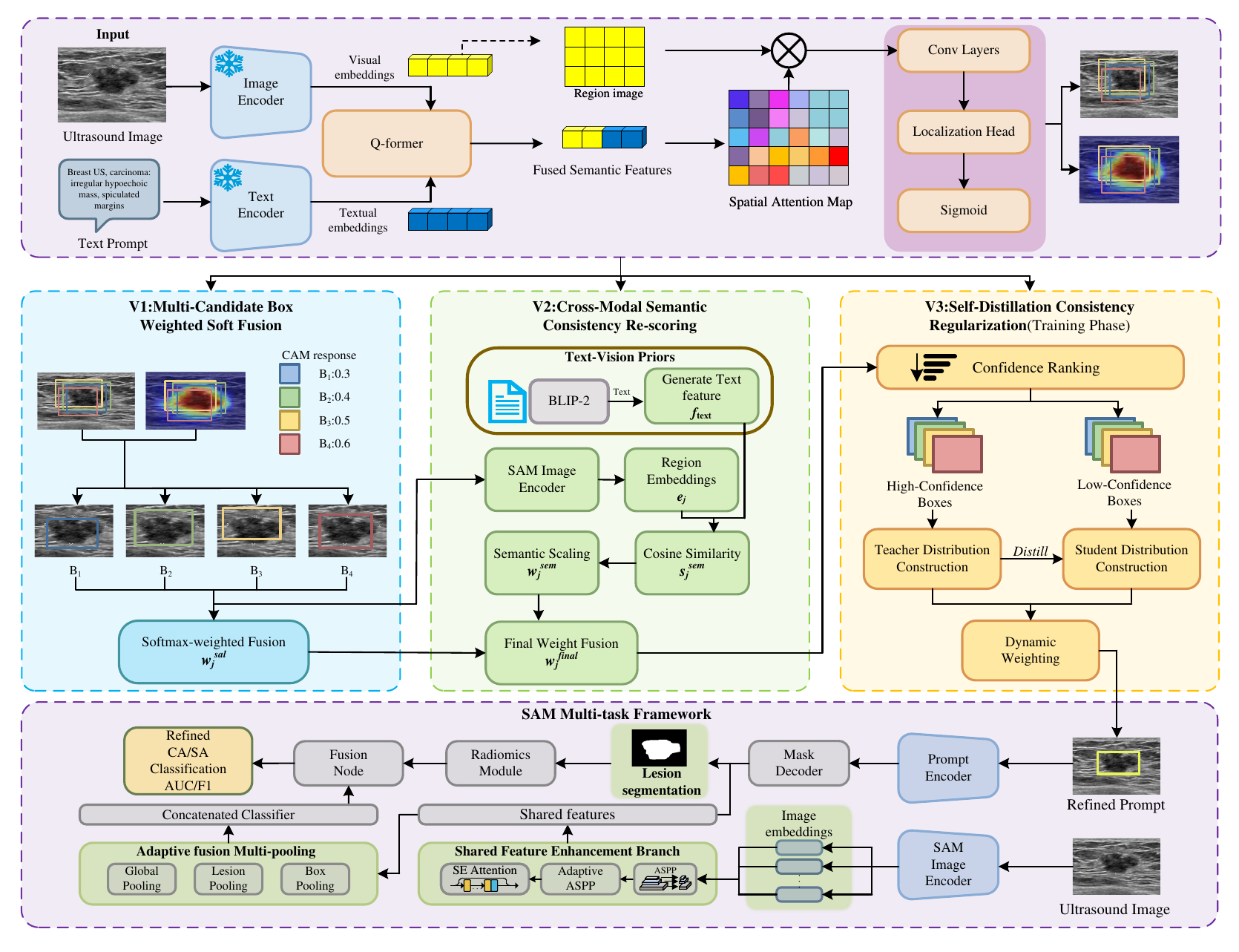}
    \caption{A Two-Stage Multimodal Prompting Framework for Breast Ultrasound Diagnosis. The framework consists of a BLIP-2-based mask-supervised semantic localization stage for automatic prompt generation with multi-candidate prompt refinement (V1-V3), and a SAM-based multi-task stage for precise lesion segmentation and visually grounded diagnosis.}
    \label{fig:framework}
\end{figure*}

Fig.~\ref{fig:framework} shows the overall workflow of the proposed framework. For an input breast ultrasound image, Stage I uses a BLIP-2-based mask-supervised localization module for automatic inference-time prompt generation to extract coarse semantic priors and convert them into candidate box prompts without manual prompt annotation at inference time. Stage II takes these candidate prompts as spatial guidance for a SAM-based multitask branch, where lesion segmentation and diagnosis are jointly performed on shared visual representations. To improve the robustness of automatically generated prompts, candidate regions are reweighted and fused according to semantic consistency and confidence cues. The final diagnosis is obtained by combining lesion-aware visual features from the SAM branch with candidate-region context and selected radiomics descriptors. Through this design, the framework establishes an evidence-driven connection from weak localization to fine-grained segmentation and downstream SA/CA classification. In our framework, cross-modal semantics is used to support prompt construction and candidate refinement, whereas the final diagnosis is performed on visually grounded lesion representations and radiomics descriptors rather than direct raw-text features.

\subsection{Language-Guided Lesion Localization and Automated Prompt Generation}
To automate prompt generation while preserving visual grounding, we employ BLIP-2 for coarse lesion localization. Importantly, linguistic signals are used exclusively to guide the generation of candidate bounding boxes and are not directly fused into the final diagnosis head. By freezing the BLIP-2 visual encoder and fine-tuning the Q-Former, we seamlessly align varying lesion regions with predefined text prompts in the target medical domain.

To prevent the model from relying on class-specific language shortcuts, we implement an asymmetric prompting strategy. During training, we use detailed BI-RADS morphological descriptions (e.g., shape, margin) to establish a robust vision-language alignment. During inference, we apply a single, fixed prompt (e.g., Breast US: abnormal lesion region) universally across all images. This strategy eliminates the need for manual annotations at test time and forces the network to ground its final diagnostic predictions purely in objective visual evidence.

Multi-Threshold Class Activation Mapping (CAM) Heatmaps and Bounding Box Generation: To convert continuous semantic activation maps into explicit spatial coordinates, we generate probability heatmaps via CAM from the text-aligned visual features. Rather than relying on a single binarization threshold, we extract candidate bounding boxes across multiple predefined thresholds. Each candidate box receives a confidence score based on its mean activation response. Finally, the boxes are ranked, and an Intersection over Union (IoU) deduplication step is applied to retain only the top-$k$ distinct bounding boxes, serving as deterministic geometric priors for the downstream network.

\subsubsection{Multi-Candidate Region Aggregation with Semantic and Confidence Constraints}
Traditional prompt-based segmentation models typically rely on a single deterministic bounding box, which is highly vulnerable to positional deviation and local noise in ultrasound imaging. To address this issue, we develop a MCRA strategy that progressively fuses candidate regions under semantic consistency and confidence-aware constraints. This design stabilizes automatically generated prompts and yields more reliable lesion-level representations for both downstream segmentation and diagnosis. The complete pipeline of this adaptive denoising and regularization mechanism is outlined in Algorithm \ref{alg:prompt_denoising}.\par
\textbf{V1: Multi-Candidate Weighted Soft Fusion}

For $K$ candidates derived from CAM multi-thresholding, SAM yields spatial masks $\mathbf{M}_j \in \mathbb{R}^{H \times W}$, classification logits $\mathbf{l}_j \in \mathbb{R}^C$, and deep embeddings $\mathbf{e}_j \in \mathbb{R}^D$. To reliably aggregate these heterogeneous representations, we normalize the initial CAM response score $s_j \in \mathbb{R}$ via temperature scaling:
\begin{equation}
    w_{j}^{\text{sal}} = \frac{\exp(s_j / \tau_{\text{sal}})}{\sum_{k=1}^K \exp(s_k / \tau_{\text{sal}})},
\end{equation}
where $\tau_{\text{sal}}$ controls the distribution sharpness. Using the semantics-calibrated weight $w_{j}^{\text{final}}$ (defined below), soft fusion is performed independently within each modality to suppress noise from low-confidence candidates:
\begin{equation}
\begin{aligned}
\mathbf{M}_{\text{fused}} &= \sum_{j=1}^K w_{j}^{\text{final}} \mathbf{M}_{j} \in \mathbb{R}^{H \times W}, \\
\mathbf{l}_{\text{fused}} &= \sum_{j=1}^K w_{j}^{\text{final}} \mathbf{l}_{j} \in \mathbb{R}^C, \\
\mathbf{e}_{\text{fused}} &= \sum_{j=1}^K w_{j}^{\text{final}} \mathbf{e}_{j} \in \mathbb{R}^D.
\end{aligned}
\end{equation}

\textbf{V2: Semantic-Guided Candidate Filtering}

Relying exclusively on spatial saliency is vulnerable to high-response artifacts (e.g., acoustic shadows). Thus, we introduce a cross-modal semantic filter to calibrate the importance of each candidate. Let $\mathbf{e}_j \in \mathbb{R}^d$ denote the SAM-encoded local region embedding of the $j$-th candidate, and $\mathbf{f}_{\text{text}} \in \mathbb{R}^d$ denote the global BLIP-2 textual representation. To ensure probabilistic consistency, we first compute the semantic matching score $s_{j}^{\text{sem}}$ via cosine similarity and normalize it into a valid distribution $w_{j}^{\text{sem}}$ via temperature scaling:
\begin{equation}
    s_{j}^{\text{sem}} = \frac{\mathbf{e}_j \cdot \mathbf{f}_{\text{text}}}{\|\mathbf{e}_j\|_2 \|\mathbf{f}_{\text{text}}\|_2}, \quad 
    w_{j}^{\text{sem}} = \frac{\exp(s_{j}^{\text{sem}} / \tau_{\text{sem}})}{\sum_{k=1}^K \exp(s_{k}^{\text{sem}} / \tau_{\text{sem}})}
\end{equation}
where $\tau_{\text{sem}}$ is a temperature hyperparameter. The final denoised weight $w_{j}^{\text{final}}$ is derived by interpolating the spatial and semantic cues:
\begin{equation}w_{j}^{\text{final}} = (1 - \alpha) w_{j}^{\text{sal}} + \alpha w_{j}^{\text{sem}}\end{equation}
where $\alpha \in [0, 1]$ modulates the cross-modal semantic contribution. This formulation ensures that $\sum_{j=1}^K w_{j}^{\text{final}} = 1$, effectively suppressing noise from candidates that lack semantic alignment. 

\textbf{V3: Self-Distillation Consistency Regularization}

In clinical ultrasound imaging, common artifacts like acoustic shadowing and speckle noise often lead to inaccurate or conflicting bounding boxes. To prevent these noisy predictions from disrupting the training process, we introduce a self-distillation consistency constraint. Intuitively, this mechanism uses reliable, high-confidence predictions to guide and correct the highly uncertain ones. Specifically, based on the calibrated weight $w_{j}^{final}$, we partition the $K$ candidates into a high-confidence subset $\mathcal{H}$ and a low-confidence subset $\mathcal{L}$. We construct a reliable teacher mask $P_{teacher} \in [0,1]^{H \times W}$ by combining the predictions from $\mathcal{H}$. A dynamically weighted Mean Squared Error (MSE) is then applied to the candidates in $\mathcal{L}$:
\begin{equation}P_{\text{teacher}} = \text{sg} \left( \sigma \left( \sum_{j \in \mathcal{H}} \tilde{w}j M_j \right) \right)
\end{equation}
\begin{equation}
\mathcal{L}{\text{cons}} = \frac{1}{|\mathcal{L}|} \sum_{j \in \mathcal{L}} (1 - w_{j}^{\text{final}})^\gamma \cdot \left| \sigma(M_j) - P_{\text{teacher}} \right|_2^2\end{equation}
where $\sigma(\cdot)$ is the sigmoid function, $\text{sg}(\cdot)$ denotes the stop-gradient operator, $\tilde{w}_j = w_j^{\text{final}} / \sum_{i \in \mathcal{H}} w_i^{\text{final}}$ is the $L_1$-normalized weight within $\mathcal{H}$, and the decay factor $\gamma > 0$ assigns aggressive penalties to highly uncertain candidates.

\begin{algorithm}[ht]
\caption{Adaptive Prompt Denoising and Consistency Regularization}
\label{alg:prompt_denoising}
\textbf{Input:} $K$ candidate boxes, CAM scores $S = \{s_j\}_{j=1}^K$, SAM predictions $\{O_j\}_{j=1}^K$ (masks $M_j$, logits $l_j$, embeddings $e_j$), text feature $f_{\text{text}}$, temperatures $\tau_{\text{sal}}$ and $\tau_{\text{sem}}$, interpolation factor $\alpha$, threshold $\theta$, decay $\gamma$.\

\textbf{Output:} Fused representations $M_{\mathrm{fused}}$, $l_{\mathrm{fused}}$, $e_{\mathrm{fused}}$, Consistency loss $\mathcal{L}_{\text{cons}}$.
\begin{algorithmic}[1]
\STATE Initialize high-confidence subset $\mathcal{H} = \emptyset$ and low-confidence subset $\mathcal{L} = \emptyset$
\FOR{$j = 1$ to $K$}
\STATE $w_j^{\text{sal}} = \frac{\exp(s_j / \tau_{\text{sal}})}{\sum_{k=1}^K \exp(s_k / \tau_{\text{sal}})}$ \hfill $\triangleright$ \textit{Compute spatial saliency via temperature scaling}
\STATE $s_j^{\text{sem}} = \frac{e_j \cdot f_{\text{text}}}{\|e_j\|_2 \|f_{\text{text}}\|_2}, \quad w_j^{\text{sem}} = \frac{\exp(s_j^{\text{sem}} / \tau_{\text{sem}})}{\sum_{k=1}^K \exp(s_k^{\text{sem}} / \tau_{\text{sem}})}$ \hfill $\triangleright$ \textit{Compute and normalize semantic alignment}
\STATE $w_j^{\text{final}} = (1 - \alpha) w_j^{\text{sal}} + \alpha w_j^{\text{sem}}$ \hfill $\triangleright$ \textit{Interpolate spatial and semantic weights}
\IF{$w_j^{\text{final}} \geq \theta$}
\STATE $\mathcal{H} = \mathcal{H} \cup \{j\}$ \hfill $\triangleright$ \textit{Assign to robust teacher subset}
\ELSE
\STATE $\mathcal{L} = \mathcal{L} \cup \{j\}$ \hfill $\triangleright$ \textit{Assign to uncertain student subset}
\ENDIF
\ENDFOR
\STATE $M_{\mathrm{fused}} = \sum_{j=1}^{K} w_j^{\mathrm{final}} M_j$ 
\STATE $l_{\mathrm{fused}} = \sum_{j=1}^{K} w_j^{\mathrm{final}} l_j$
\STATE $e_{\mathrm{fused}} = \sum_{j=1}^{K} w_j^{\mathrm{final}} e_j$ \hfill $\triangleright$ \textit{Derive final representations via adaptive soft fusion}
\STATE $\tilde{w}_j = \frac{w_j^{\text{final}}}{\sum_{i \in \mathcal{H}} w_i^{\text{final}}}, \quad \forall j \in \mathcal{H}$ \hfill $\triangleright$ \textit{$L_1$ re-normalization for high-confidence weights}
\STATE $P_{\text{teacher}} = \text{sg} \left( \sigma \left( \sum_{j \in \mathcal{H}} \tilde{w}_j M_j \right) \right)$ \hfill $\triangleright$ \textit{Construct robust teacher target (stop-gradient)}
\STATE $\mathcal{L}_{\text{cons}} = \frac{1}{|\mathcal{L}|} \sum_{j \in \mathcal{L}} (1 - w_j^{\text{final}})^\gamma \cdot \left\| \sigma(M_j) - P_{\text{teacher}} \right\|_2^2$ \hfill $\triangleright$ \textit{Dynamically weighted MSE penalty}
\RETURN $M_{\mathrm{fused}}$, $l_{\mathrm{fused}}$, $e_{\mathrm{fused}}$, $\mathcal{L}_{\text{cons}}$
\end{algorithmic}
\end{algorithm}

\subsubsection{Joint Optimization for Cross-Modal Alignment and Localization}
Rather than isolating modalities early on, Stage I explicitly learns cross-modal alignment to facilitate weak localization and robust prompt construction. To this end, we define a joint objective that functionally mirrors our semantics-assisted prompting and multi-candidate generation strategies:
\begin{equation}
\mathcal{L}_{Stage1} = \lambda_{cls}\mathcal{L}_{cls} + \lambda_{loc}\mathcal{L}_{loc} + \lambda_{text}\mathcal{L}_{text\_aux}
\end{equation}
where $\lambda_{cls}$, $\lambda_{loc}$, and $\lambda_{text}$ are weights utilized to balance these objectives.

Specifically, $\mathcal{L}_{loc}$ operationalizes the foundational premise of the V1 module. By comparing the CAM logits against the ground-truth segmentation masks using a combination of Binary Cross-Entropy (BCE) and soft Dice loss, this term ensures that the predicted activation map accurately covers the lesion region. This dense localization supervision ensures that the subsequent multi-thresholding process extracts valid, geometrically sound candidate bounding boxes. The term $\mathcal{L}_{cls}$ serves as the primary supervision signal for benign-versus-malignant discrimination at the image level, implemented as a class-balanced focal loss on the classification logits.

Furthermore, $\mathcal{L}_{text\_aux}$ provides the optimization objective for our asymmetric text-prompting strategy and V2 semantic re-scoring mechanism. This description-guided auxiliary loss explicitly forces the visual encoder to align with the fine-grained BI-RADS semantic descriptors (e.g., shape, margin) provided during training. By establishing this robust joint embedding space, this early cross-modal alignment ensures that the BLIP-2 text features used for computing cosine similarity in V2 contain rich clinical semantics. Consequently, the automated prompt construction is deeply informed by these semantics, enabling the framework to effectively filter out spurious visual artifacts such as acoustic shadows without relying on manual annotations at inference time.

\subsection{SAM-Based Multi-Task Segmentation and Classification Network}
To couple fine-grained segmentation with reliable diagnosis, Stage II introduces a SAM-based multi-task architecture built upon a shared visual backbone. Given an ultrasound image and the automatically generated candidate boxes, SAM's image and prompt encoders collaborate to output precise lesion masks via the mask decoder. Instead of treating diagnosis as an isolated task, we cascade these predicted masks as explicit spatial priors into a parallel classification branch, which directly extracts the shared high-dimensional visual features from the SAM encoder to synergistically model both global lesion characteristics and local morphological details. Crucially, this downstream diagnostic head relies exclusively on these objective, lesion-level visual representations, purposefully excluding raw textual embeddings to ensure the final benign/malignant prediction remains strictly evidence-driven and visually grounded.

\subsubsection{Multi-Scale Feature Adaptive Aggregation}
To address the pronounced scale and morphological variations of ultrasound lesions, we introduce a multi-scale contextual aggregation scheme following the SAM Image Encoder.

Given the visual features $X \in \mathbb{R}^{C \times H \times W}$, we first apply a Squeeze-and-Excitation (SE) mechanism to adaptively amplify lesion-sensitive channels and suppress background noise. The recalibrated feature is derived as $\tilde{X} = s \odot X$, where the channel descriptor $s$ is computed by:
\begin{equation}s = \sigma\left(W_2 \delta(W_1 (\text{GAP}(X)))\right)\end{equation}
with $\text{GAP}(\cdot)$ denoting global average pooling, $\delta(\cdot)$ the ReLU activation, and $\sigma(\cdot)$ the Sigmoid function.

To capture diverse geometric contexts, $\tilde{X}$ is subsequently processed via parallel Atrous Spatial Pyramid Pooling (ASPP) with dilation rates $r \in \{1, 6, 12, 18\}$, yielding a multi-scale representation $F_{\text{ASPP}}$. To prevent the dilution of localized boundary details typically caused by naive element-wise addition, we employ an adaptive spatial gating strategy to fuse the representations:
\begin{equation}G = \sigma\left(\text{Conv}{1 \times 1}([\tilde{X}; F{\text{ASPP}}])\right)\end{equation}\begin{equation}X_{\text{fused}} = \tilde{X} + G \odot (F_{\text{ASPP}} - \tilde{X})\end{equation}
where $[\cdot;\cdot]$ denotes channel concatenation. This soft residual injection dynamically assimilates broad anatomical context while preserving the fine-grained visual boundaries essential for precise delineation.

\subsubsection{Multi-Pooling Enhanced Classifier}
Although the SAM Mask Decoder is capable of outputting high-resolution segmentation logits, the benign/malignant classification task imposes more stringent representation requirements on the infiltration characteristics of the lesion and its surrounding tissues. In the classification branch, we discard the traditional global pooling paradigm and construct a multi-dimensional pooling feature concatenation strategy.

Let $M_{\text{lesion}}, M_{\text{box}} \in \{0,1\}^{1 \times H \times W}$ represent the predicted lesion mask and bounding box mask, respectively. We utilize these as spatial priors to perform precise region-aware pooling on the shared feature $X_{\text{fused}}$, thereby extracting the local heterogeneous features of the lesion core and proximal neighborhood:
\begin{equation}
F_{\text{lesion}} = \frac{\sum_{i,j} X_{\text{fused}}(i,j) \cdot M_{\text{lesion}}(i,j)}{\sum_{i,j} M_{\text{lesion}}(i,j) + \epsilon}
\end{equation}
\begin{equation}
F_{\text{box}} = \frac{\sum_{i,j} X_{\text{fused}}(i,j) \cdot M_{\text{box}}(i,j)}{\sum_{i,j} M_{\text{box}}(i,j) + \epsilon}
\end{equation}
where $\epsilon$ is a small constant to prevent division by zero. Subsequently, these two region-level features are concatenated along the channel dimension with the Global Average Pooling feature ($F_{\text{GAP}}$) and Global Max Pooling feature ($F_{\text{GMP}}$), which represent the macroscopic context:
\begin{equation}
F_{\text{pool}} = [F_{\text{lesion}}; F_{\text{box}}; F_{\text{GAP}}; F_{\text{GMP}}]
\end{equation}

The final classification logits are predicted by a Multi-Layer Perceptron (MLP): $Y_{\text{cls}} = \text{MLP}(F_{\text{pool}})$. This multi-dimensional fusion strategy compels the classification network not only to focus deeply on the internal echoic structural distribution of the lesion but also to sensitively capture critical surrounding tissue infiltration features essential for ultrasound clinical diagnosis (e.g., spiculated margins, posterior acoustic shadowing), thereby significantly enhancing the robustness of the classification boundary.

\subsection{Interpretable Dual-Branch Decision Fusion}
Conventional deep learning models are often criticized as ``black boxes'' due to their lack of clinical transparency. To address this limitation and bridge the gap between implicit deep representations and explicit physical priors, we propose an interpretable dual-branch decision fusion module. During the forward pass, we leverage the predicted lesion mask as a precise spatial prior to extract a high-dimensional radiomics feature pool comprising over 500 dimensions. These raw features provide a comprehensive quantification of lesion heterogeneity, primarily encompassing morphological features (e.g., geometric contours), first-order statistics (e.g., echo intensity distribution), and high-order texture features (e.g., GLCM, GLRLM, GLSZM).

To ensure feature representation efficiency and prevent overfitting, a cascaded selection strategy employing mRMR and LASSO is implemented to retain the 63 most discriminative radiomics features. This selection process is performed exclusively on the training split, the resulting feature subset is then fixed and consistently applied to the validation and test sets. These refined core features are then processed by a dedicated Radiomics Head to generate radiomics-specific logits. The final classification probability is determined by the synergistic fusion of the SAM-enhanced visual classification head and the explicit radiomics head. The fusion process is formulated as:
\begin{equation}
Y_{\text{final}} = Y_{\text{SAM}} + \alpha_{\text{ext}} Y_{\text{ext\_rad}}
\end{equation}
where $Y_{\text{SAM}}$ denotes the classification output derived from the deep implicit representations of the shared backbone, $Y_{\text{ext\_rad}}$ represents the diagnostic prediction from the explicitly selected external radiomic features, and $\alpha_{\text{ext}}$ serves as the fusion balance weight. This dual-branch mechanism compellingly integrates complementary information sources, anchoring the diagnostic reasoning with robust, physically interpretable constraints and transcending the inherent opacity of purely data-driven visual models.

\section{Experiments}

\begin{figure*}[t] 
    \centering
    \includegraphics[width=\textwidth]{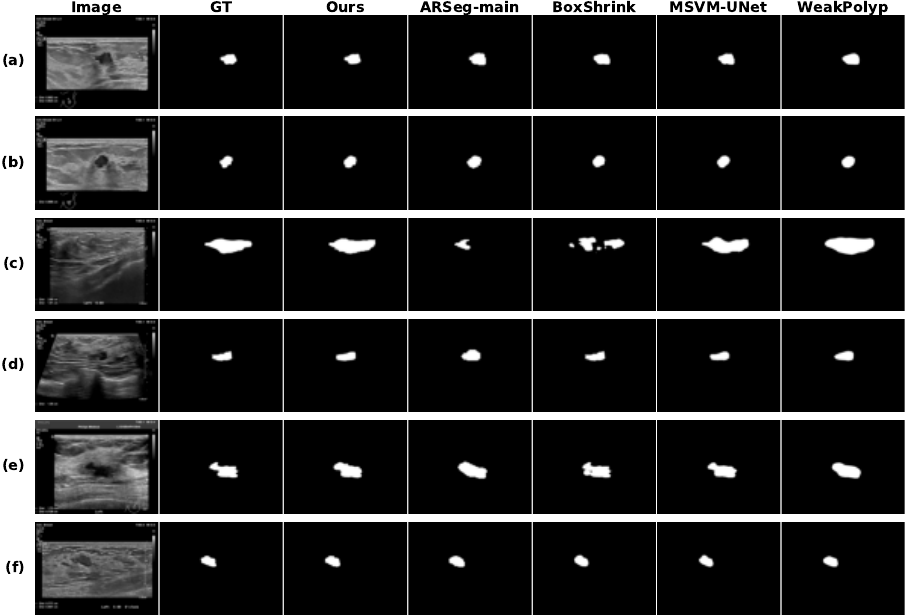}
    \caption{Qualitative comparison of lesion segmentation performance on challenging ultrasound cases. The selected samples (rows a--f) are characterized by severe background interference, indistinct lesion-tissue transitions, and ambiguous acoustic shadows. We compare our Rad-VLSM against Ground Truth (GT) annotations and four state-of-the-art baselines. While competing methods are highly susceptible to spurious artifacts—often resulting in severe under-segmentation, fragmented predictions, or failure to capture complex morphological protrusions —our proposed Rad-VLSM demonstrates strong resilience, accurately delineating fine-grained contours that closely mirror the GT.}
    \label{fig:main_architecture}
\end{figure*}

\subsection{Datasets}
\subsubsection{Clinical Breast Ultrasound Dataset} In this study, we use a private clinical image dataset provided by Peking Union Medical College Hospital for algorithm evaluation. The dataset comprises 559 aligned ultrasound images, categorizing lesions into Cancer (CA, 291 images) and Sclerosing Adenosis (SA, 268 images). Each image is accompanied by a precise binary segmentation mask delineating the lesion region. The dataset is randomly partitioned into training and validation sets at an 8:2 ratio.
\subsubsection{ISIC 2016\cite{gutman2016skin} and ISIC 2018 Datasets\cite{codella2019skin}}To evaluate our method on dermoscopic imagery, we utilized the ISIC 2016 and ISIC 2018 datasets. The ISIC 2016 dataset includes 900 training and 379 testing images, while ISIC 2018 provides a larger collection of 2,594 images.
\subsubsection{BUSI Dataset\cite{al2020dataset}}The Breast Ultrasound Images (BUSI) dataset consists of 697 images (487 benign and 210 malignant). Original DICOM images were converted to PNG and transformed to grayscale prior to our standardized resizing protocol.
\subsubsection{Colonic Polyp Dataset}We curated a comprehensive dataset encompassing 3,784 endoscopic images by aggregating samples from five benchmarks: CVC-ClinicDB\cite{bernal2015wm}, CVC-ColonDB\cite{tajbakhsh2015automated}, ETIS-LaribPolypDB\cite{silva2014toward}, Kvasir\cite{jha2019kvasir}, and PolypGen\cite{ali2023multi,ali2024assessing,ali2021deep}. The training/validation split followed an 8:2 ratio.
\subsubsection{MRI Brain Tumor Dataset\cite{buda2019association}}We incorporated the LGG Segmentation Dataset from The Cancer Imaging Archive, comprising 3,929 brain MRI images. Experimental settings remain consistent with the Colonic Polyp dataset.

\subsection{Implementation Details and Evaluation Metrics}
\subsubsection{Implementation Details}
All experiments were implemented using the PyTorch framework and accelerated on a single NVIDIA RTX 4090 GPU. To reduce memory footprint and expedite the training process, Automatic Mixed Precision (AMP) was universally enabled. We adopted a robust two-stage training paradigm. In Stage 1, the weak localization network, integrating BLIP-2, is trained for 30 epochs to generate CAMs, which subsequently yield coarse bounding box prompts. In Stage 2, the SAM is fine-tuned for 50 epochs, utilizing the CAM-generated bounding boxes as explicit prompts to guide precise lesion segmentation. Notably, for the breast-related datasets (Custom Breast and BUSI), a cascaded execution of segmentation followed by binary classification is performed during this second stage. A consistent batch size of 4 was maintained across all stages and datasets. Regarding input spatial dimensions, Stage 1 utilized domain-specific resolutions: 1024× 1024 for the clinical breast, BUSI, and ISIC datasets, and 512 × 512 for the Colonic Polyp and MRI Brain Tumor datasets. During Stage 2, the input resolution for all datasets was uniformly standardized to 1024 × 1024. Optimization Strategy: For the optimization scheme, both training stages employed the AdamW optimizer with an initial learning rate of $1 \times 10^{-4}$ and a weight decay coefficient of $1 \times 10^{-4}$. To stabilize the early training phase and ensure smooth convergence, the learning rate was governed by a Cosine Annealing scheduler, which included a linear warmup period of 5 epochs and gradually decayed to a minimum learning rate of $1 \times 10^{-6}$.

\subsubsection{Evaluation Metrics}
To rigorously assess the performance of our proposed framework, we employ a comprehensive suite of evaluation metrics encompassing both lesion segmentation and binary classification tasks. All metrics are continuously monitored and recorded on the held-out validation set at the end of each training epoch to track the model's generalization capability.

Segmentation Metrics: For the segmentation task, we adopt two standard region-based spatial overlap indices to quantify the pixel-level agreement between the predicted masks and the ground-truth annotations: the mean Dice Similarity Coefficient (mDSC) and the mean Intersection over Union (mIoU).

Classification Metrics: To evaluate the binary classification performance, we compute the Area Under the Receiver Operating Characteristic Curve (ROC-AUC) to measure the model's overall discriminative capacity across various thresholds. Furthermore, by applying a standard fixed decision threshold of 0.5 to the predicted probabilities, we calculate a holistic set of complementary metrics: Accuracy (Acc), Precision, Sensitivity (Sens), Specificity (Spec), and the harmonic mean of precision and recall, the F1-score. This combination ensures a balanced and robust assessment of the model's diagnostic reliability.

\subsection{Comparisons With Other Methods}
\textbf{Segmentation}
\subsubsection{Results on Our Datasets}\par

We categorized 13 state-of-the-art (SOTA) baselines into three paradigms: Mainstream encoder-decoder (UNet~\cite{ronneberger2015u}, Att-UNet); Transformer-based (Swin-UNet, TransUNet~\cite{chen2024transunet}, EMCAD, PVT-Cascade); and Weakly-supervised/Foundation models (SimTxtSeg~\cite{xie2024simtxtseg}, BoxShrink~\cite{groger2022boxshrink}, MedCLIP-SAMv2).\par
\newcolumntype{Y}{>{\centering\arraybackslash}X}

\begin{table}[htbp]
    \centering
    \renewcommand{\arraystretch}{1.2} 
    \caption{Quantitative comparison of lesion segmentation performance on the clinical breast ultrasound dataset. Our method is compared against 11 state-of-the-art baselines. The best results are highlighted in bold.}
    \label{tab:segmentation_results}
    
    \begin{tabularx}{\columnwidth}{l YY} 
        \toprule
        \textbf{Models} & \textbf{mDSC} $\uparrow$ & \textbf{mIoU} $\uparrow$ \\
        \midrule
        UNet            & 0.6545 & 0.6035 \\
        Att-UNet        & 0.6558 & 0.6054 \\
        \midrule
        Swin-UNet       & 0.6629 & 0.6035 \\
        TransUNet       & 0.7198 & 0.6551 \\
        EMCAD           & 0.7462 & 0.6919 \\
        PVT-Cascade     & 0.7324 & 0.6715 \\
        \midrule
        SimTxtSeg       & 0.8403 & 0.7444 \\
        ARSeg           & 0.8088 & 0.6875 \\
        WeakPolyp       & 0.8512 & 0.7515 \\
        S2ME            & 0.6115 & 0.4514 \\
        BoxShrink       & 0.8735 & 0.8039 \\
        MedCLIP-SAMv2   & 0.8865 & 0.8169     \\ 
        \midrule
        \textbf{Ours (Rad-VLSM)} & \textbf{0.9217} & \textbf{0.8579} \\
        \bottomrule
    \end{tabularx}
\end{table}

As summarized in Table \ref{tab:segmentation_results}, our proposed method consistently outperforms competitors. Compared to fully supervised networks like TransUNet, our approach achieves superior mDSC, demonstrating robust boundary delineation without relying on manual prompt annotations at inference time. Our two-stage paradigm exhibits a significant margin over prompt-driven SAM variants like BoxShrink, validating that our CAM-guided strategy effectively bridges the semantic gap.\par

Figure \ref{fig:main_architecture} illustrates segmentation results on challenging ultrasound cases characterized by acoustic shadows and indistinct lesion transitions (e.g., rows c and e). While baseline methods such as BoxShrink suffer from severe under-segmentation or fragmented regional predictions due to susceptibility to these artifacts, Rad-VLSM maintains topological integrity. This spatial stability is primarily attributed to the V1-V3 multi-candidate prompt aggregation mechanism, which explicitly suppresses non-lesion noise and aligns highly uncertain candidate regions with the morphological consensus of robust predictions.

\begin{table}[htbp]
\centering
\caption{Quantitative comparison of lesion segmentation performance on ISIC 2016 and ISIC 2018 datasets. The best results are highlighted in bold.}
\label{tab:isic_comparison}
\renewcommand{\arraystretch}{1.1} 
\setlength{\tabcolsep}{1.2pt}      

\begin{tabularx}{\columnwidth}{@{\extracolsep{\fill}} l *{4}{S[table-format=1.4]} @{}}
\toprule
\multirow{2}{*}{\textbf{Models}} & \multicolumn{2}{c}{\textbf{ISIC 2016}} & \multicolumn{2}{c}{\textbf{ISIC 2018}} \\ 
\cmidrule(lr){2-3} \cmidrule(lr){4-5}
& {\textbf{mDSC} $\uparrow$} & {\textbf{mIoU} $\uparrow$} & {\textbf{mDSC} $\uparrow$} & {\textbf{mIoU} $\uparrow$} \\ 
\midrule
UNet              & 0.8984 & 0.8315 & 0.8386 & 0.7507 \\
Att-UNet    & 0.8743 & 0.7970 & 0.8388 & 0.7594 \\
DeepLabv3+        & 0.9012 & 0.8435 & 0.8938 & 0.8249 \\
UNeXt             & 0.9103 & 0.8397 & 0.8981 & 0.8164 \\
nnUNet            & 0.9045 & 0.8452 & 0.9022 & 0.8376 \\
MALUNet           & 0.8812 & 0.8025 & 0.8672 & 0.7863 \\
EGE-UNet          & 0.9019 & 0.8188 & 0.8783 & 0.7956 \\
Swin-Unet         & 0.9012 & 0.8321 & 0.8972 & 0.8290 \\
CTO               & 0.9189 & 0.8518 & 0.9120 & 0.8450 \\
UCTransNet        & 0.9058 & 0.8414 & 0.8671 & 0.7843 \\
STM-UNet          & 0.9094 & 0.8463 & 0.8751 & 0.7984 \\
TransUNet         & 0.9131 & 0.8496 & 0.8875 & 0.8178 \\
SegFormer         & 0.9131 & 0.8496 & 0.8875 & 0.8178 \\
FCBFormer         & 0.9262 & 0.8712 & 0.9129 & 0.8487 \\
H2Former          & 0.9241 & 0.8645 & 0.9117 & 0.8435 \\
VM-UNet           & 0.9007 & 0.8283 & 0.8971 & 0.8135 \\
MiM-ISTD          & 0.9225 & 0.8628 & 0.9062 & 0.8379 \\
BSBP-RWKV         & 0.9351 & 0.8847 & 0.9148 & 0.8611 \\
Zig-RiR           & 0.9287 & 0.8818 & 0.9170 & 0.8689 \\
Med-FastSAM       & {--}   & {--}   & 0.8784 & 0.8038 \\
SAMed             & {--}   & {--}   & 0.8693 & 0.7898 \\
\midrule
\textbf{Ours (Rad-VLSM)} & \textbf{0.9368} & \textbf{0.8843} & \textbf{0.9331} & \textbf{0.8786} \\
\bottomrule
\end{tabularx}
\end{table}

\begin{table*}[t]
    \centering
    \caption{Comprehensive comparison of joint segmentation and classification performance. Segmentation metrics are reported as coefficients (0--1), while diagnostic metrics are reported in percentage (\%). The best results are highlighted in \textbf{bold}.}
    \label{tab:unified_multitask}
    \renewcommand{\arraystretch}{1.2} 
    \small
    \begin{tabularx}{\textwidth}{@{\extracolsep{\fill}} l cc cccc @{}}
        \toprule
        \multirow{2}{*}{\textbf{Methods}} & \multicolumn{2}{c}{\textbf{Segmentation}} & \multicolumn{4}{c}{\textbf{Diagnostic Classification (\%)}} \\ 
        \cmidrule(lr){2-3} \cmidrule(lr){4-7}
        & \textbf{mDSC} $\uparrow$ & \textbf{mIoU} $\uparrow$ & \textbf{ACC} $\uparrow$ & \textbf{Rec.} $\uparrow$ & \textbf{Prec.} $\uparrow$ & \textbf{F1} $\uparrow$ \\ 
        \midrule
        \multicolumn{7}{c}{\textit{Standard \& Attentional Architectures}} \\
        \midrule
        U-Net            & 0.7850 & 0.6461 & 96.34 & 73.86 & 93.63 & 82.58 \\
        SegNet           & 0.6557 & 0.5895 & 82.13 & 69.44 & 91.26 & 78.87 \\
        U-Net++          & 0.7906 & 0.6537 & 96.07 & 76.21 & 94.90 & 84.53 \\
        Unext            & 0.8059 & 0.6742 & 95.92 & 80.95 & 97.58 & 88.49 \\
        CUNet            & 0.5978 & 0.5278 & 89.72 & 72.52 & 93.57 & 81.67 \\
        CA-Net           & 0.7977 & 0.6634 & 96.54 & 78.67 & 97.16 & 86.94 \\
        U-NetAFF         & 0.7881 & 0.6503 & 95.85 & 79.29 & 95.13 & 86.49 \\
        U-Net-iAFF       & 0.7975 & 0.6632 & 95.88 & 83.40 & 94.62 & 88.65 \\
        TransUnet        & 0.7671 & 0.6615 & 93.46 & 75.87 & 92.86 & 83.49 \\
        SwinUnet         & 0.8053 & 0.6892 & 96.43 & 81.71 & 96.58 & 88.52 \\
        DCSAU            & 0.7403 & 0.5877 & 94.88 & 75.10 & 93.45 & 83.27 \\
        AHF-UNet (UAFF)  & 0.8195 & 0.6942 & 96.49 & 82.01 & 96.53 & 88.68 \\
        AHF-UNet (CLSCA) & 0.8243 & 0.7012 & 96.63 & 81.91 & 97.13 & 88.87 \\
        \midrule
        \multicolumn{7}{c}{\textit{Dedicated Multi-Task Paradigms}} \\
        \midrule
        Bruno et al.            & 0.7690 & 0.6250 & 95.20 & 92.40 & 96.60 & 94.40 \\
        Mask R-CNN              & 0.7620 & 0.6910 & 80.90 & 74.40 & 76.80 & 75.40 \\
        Multi-task Transformers & 0.6170 & 0.5120 & 89.80 & 65.40 & 94.40 & 77.30 \\
        \midrule
        \textbf{Ours (Rad-VLSM)} & \textbf{0.9075} & \textbf{0.8351} & \textbf{97.67} & \textbf{97.62} & \textbf{97.69} & \textbf{97.65} \\
        \bottomrule
    \end{tabularx}
\end{table*}
\subsubsection{Results on Other Datasets}
To demonstrate cross-modal generalizability, we extended evaluation to MRI, endoscopy, and dermoscopy benchmarks. As reported in Table \ref{tab:cross_modal_final}, our method achieved a highly competitive mDSC of 0.9258 on the MRI Brain Tumor dataset, outperforming the second-best method (PraNet) by over 10\%. Stable performance across diverse morphologies (Skin lesions, Polyps) underscores the universal applicability of our CAM-guided prompt generation.\par
Furthermore, the robustness of Rad-VLSM is further validated on dermatoscopic imagery, as detailed in Table \ref{tab:isic_comparison}. Skin lesion segmentation is notoriously challenging due to inherently fuzzy lesion-skin transitions, highly variable lighting, and pervasive artifacts (e.g., hair occlusions and ruler markings). Despite these extreme obstacles, our method establishes new state-of-the-art performance on the large-scale ISIC 2018 benchmark, achieving an mDSC of 0.9331 and an mIoU of 0.8786. This outperforms recent advanced architectures specifically optimized for boundary refinement, such as Zig-RiR (mDSC 0.9170) and BSBP-RWKV (mDSC 0.9148). A similarly dominant trend is observed on the ISIC 2016 dataset, where Rad-VLSM yields a leading mDSC of 0.9368. These consistent multi-dataset breakthroughs consistently demonstrate that our two-stage integration of semantic-aware prompt generation and high-fidelity SAM decoding successfully mitigates boundary ambiguity across highly diverse clinical modalities.

\begin{table}[htbp]
    \centering
    \renewcommand{\arraystretch}{1.2} 
    \caption{Performance comparison on MRI Brain Tumor and Colonic Polyp datasets. The best results are highlighted in bold.}
    \label{tab:cross_modal_final}
    
    \begin{tabularx}{\columnwidth}{@{\extracolsep{\fill}} l *{4}{S[table-format=1.4]} @{}}
        \toprule
        \multirow{2}{*}{\textbf{Models}} & \multicolumn{2}{c}{\textbf{MRI Brain Tumor}} & \multicolumn{2}{c}{\textbf{Colonic Polyp}} \\ 
        \cmidrule(lr){2-3} \cmidrule(lr){4-5}
        & {\textbf{mDSC} $\uparrow$} & {\textbf{mIoU} $\uparrow$} & {\textbf{mDSC} $\uparrow$} & {\textbf{mIoU} $\uparrow$} \\ 
        \midrule
        SimTxtSeg          & 0.8174 & 0.7134 & 0.8824 & 0.8247 \\
        BoxShrink          & 0.6636 & 0.5702 & 0.7821 & 0.6422 \\
        BoxPolyp           & 0.7764 & 0.6740 & 0.8686 & 0.7911 \\
        S2ME               & 0.2666 & 0.1538 & 0.6633 & 0.4962 \\
        WeakPolyp          & 0.7482 & 0.6343 & 0.8561 & 0.7940 \\
        ResUNet            & 0.7127 & 0.5842 & 0.8260 & 0.7531 \\
        PraNet             & 0.8249 & 0.7414 & 0.8730 & 0.8132 \\
        Ariadne's Thread   & 0.8120 & 0.7155 & 0.8714 & 0.8065 \\
        \midrule
        \textbf{Ours}      & \textbf{0.9258} & \textbf{0.8653} & \textbf{0.9294} & \textbf{0.8763} \\
        \bottomrule
        \addlinespace[2pt]
    \end{tabularx}
\end{table}

\textbf{Classification}\par

\begin{table}[htbp]
    \centering
    \caption{Quantitative comparison of binary classification performance on the clinical breast ultrasound dataset. All metrics are reported in percentage (\%). The best results are highlighted in bold.}
    \label{tab:classification_comparison}
    \renewcommand{\arraystretch}{1.2} 
    
    \begin{tabularx}{\columnwidth}{@{\extracolsep{\fill}} l S[table-format=2.2] S[table-format=2.2] S[table-format=2.2] S[table-format=2.2] S[table-format=2.2] @{}}
        \toprule
        \textbf{Models} & {\textbf{ACC}} & {\textbf{AUC}} & {\textbf{Sens}} & {\textbf{Spec}} & {\textbf{F1}} \\
        \midrule
        MambaMIC            & 75.90 & 82.27 & 85.00 & 67.44 & 75.76 \\
        ADSR-net            & 79.46 & 87.93 & 77.78 & 81.03 & 78.50 \\
        MIAFEx              & 82.30 & 87.82 & 82.24 & 82.24 & 82.29 \\
        MedMamba            & 74.70 & 76.28 & 74.40 & 75.00 & 74.70 \\
        DiffMIC             & 76.79 & 86.59 & 68.52 & 84.48 & 74.00 \\
        \midrule
        ResNet50            & 71.08 & 79.59 & 65.12 & 77.50 & 71.01 \\
        Densenet            & 75.90 & 80.17 & 83.72 & 67.50 & 75.71 \\
        ConvNext-Tiny       & 57.83 & 58.49 & 60.47 & 55.00 & 57.81 \\
        ViT                 & 73.49 & 72.13 & 51.16 & 97.50 & 73.78 \\
        PVT-Small           & 59.04 & 59.53 & 81.40 & 35.00 & 56.63 \\
        UniFormer-Small     & 68.67 & 71.45 & 72.09 & 65.00 & 68.63 \\
        Vim-Tiny            & 57.83 & 63.43 & 86.05 & 27.50 & 53.77 \\
        \midrule
        \textbf{Ours}       & \textbf{94.07} & \textbf{97.55} & \textbf{98.41} & \textbf{89.09} & \textbf{94.66} \\
        \bottomrule
    \end{tabularx}
\end{table}

\newcommand{\cmark}{\checkmark}
\newcommand{\xmark}{$\times$}
\begin{table*}[t]
    \centering
    \caption{Contribution Attribution of Architecture Components. The ablation evaluates the impact of the interpretable dual-branch fusion (SAM Base and Radiomics) alongside the multi-task feature synergy modules (SE, Adaptive ASPP, Multi-pooling). Best results are highlighted in \textbf{bold}.}
    \label{tab:unified_ablation}
    \renewcommand{\arraystretch}{1.2}
    \begin{tabular*}{\textwidth}{@{\extracolsep{\fill}} ccccc ccccc @{}}
        \toprule
        \multicolumn{5}{c}{\textbf{Architecture Components}} & \multicolumn{5}{c}{\textbf{Diagnostic Metrics (\%)}} \\
        \cmidrule(r){1-5} \cmidrule(l){6-10}
        \textbf{SAM Base} & \textbf{Radiomics} & \textbf{SE} & \textbf{Ada-ASPP} & \textbf{Multi-Pool} & \textbf{ACC $\uparrow$} & \textbf{AUC $\uparrow$} & \textbf{Sens. $\uparrow$} & \textbf{Spec. $\uparrow$} & \textbf{F1 $\uparrow$} \\
        \midrule
        \xmark & \cmark & \xmark & \xmark & \xmark & 86.44 & 94.70 & 87.30 & 85.45 & 87.30 \\
        \cmark & \cmark & \xmark & \xmark & \xmark & 83.90 & 92.99 & 82.54 & 85.45 & 84.55 \\
        \cmark & \xmark & \cmark & \cmark & \cmark & 89.83 & 92.84 & 88.89 & \textbf{90.91} & 90.32 \\
        \cmark & \cmark & \xmark & \cmark & \cmark & 91.53 & 97.06 & 95.24 & 87.27 & 92.31 \\
        \cmark & \cmark & \cmark & \xmark & \cmark & 88.98 & 95.35 & 90.48 & 87.27 & 89.76 \\
        \cmark & \cmark & \cmark & \cmark & \xmark & 85.59 & 93.30 & 84.13 & 87.27 & 86.18 \\
        \midrule
        \cmark & \cmark & \cmark & \cmark & \cmark & \textbf{94.07} & \textbf{97.55} & \textbf{98.41} & 89.09 & \textbf{94.66} \\
        \bottomrule
    \end{tabular*}
\end{table*}

\begin{table*}[t]
    \centering
    \caption{Quantitative evaluation of different bounding box strategies and variants (V1, V2, V3) on the dataset. The best results for each metric are highlighted in \textbf{bold}.}
    \label{tab:multi_task_formatted}
    
    \begin{tabular*}{\textwidth}{@{\extracolsep{\fill}} cc ccc ccccccc @{}}
        \toprule
        \textbf{GT-box} & \textbf{CAM-box} & \textbf{V1} & \textbf{V2} & \textbf{V3} & \textbf{mDSC} & \textbf{mIoU} & \textbf{ACC} & \textbf{AUC} & \textbf{Sens} & \textbf{Spec} & \textbf{F1} \\
        \midrule
        
        \cmark & \xmark & \xmark & \xmark & \xmark & \textbf{0.9270} & \textbf{0.8671} & 77.97 & 86.06 & 77.78 & 78.18 & 79.03 \\
        \midrule
        
        \multirow{5}{*}{\xmark} & \multirow{5}{*}{\cmark} 
        & \xmark & \xmark & \xmark & 0.9224 & 0.8589 & 83.90 & 91.34 & 87.30 & 80.00 & 85.27 \\
        & & \cmark & \xmark & \xmark & 0.9232 & 0.8600 & 84.75 & 91.11 & 88.89 & 80.00 & 86.15 \\
        & & \cmark & \cmark & \xmark & 0.9175 & 0.8515 & 86.44 & 90.51 & \textbf{92.06} & 80.00 & 87.88 \\
        & & \cmark & \xmark & \cmark & 0.9127 & 0.8431 & 85.59 & 92.47 & 80.95 & \textbf{90.91} & 85.71 \\
        & & \cmark & \cmark & \cmark & 0.9158 & 0.8465 & \textbf{89.83} & \textbf{92.84} & 88.89 & \textbf{90.91} & \textbf{90.32} \\
        
        \bottomrule
    \end{tabular*}
\end{table*}

We evaluate the diagnostic capability of our framework by conducting binary classification on the clinical breast ultrasound dataset. We benchmark our method against twelve diverse architectures, encompassing standard convolutional and Transformer-based backbones (e.g., ResNet50, DenseNet, ConvNext-Tiny, ViT, PVT-Small), as well as recent state-of-the-art models specifically tailored for medical imaging and advanced sequence modeling (e.g., MambaMIC~\cite{zou2024microscopic}, MedMamba~\cite{yue2024medmamba}, DiffMIC~\cite{yang2023diffmic}, MIAFEx~\cite{ramos2025miafex}, and ADSR-net~\cite{zhang2025adaptive}). 

As shown in Table~\ref{tab:classification_comparison}, Rad-VLSM outperforms the strongest baseline (MIAFEx) by a significant margin of 11.77\% in Accuracy and 9.73\% in AUC. Furthermore, while several baseline models exhibit heavily skewed predictions---for instance, ViT sacrifices Sensitivity (51.16\%) for Specificity (97.50\%)---our framework achieves a balanced diagnostic profile (Sensitivity: 98.41\%, Specificity: 89.09\%). This balance highlights the effectiveness of the interpretable dual-branch decision fusion in capturing both global macroscopic context and fine-grained perilesional evidence.

\textbf{Segmentation and Classification}

To evaluate the unified efficacy of the proposed framework, we benchmark Rad-VLSM against a comprehensive set of joint segmentation and classification paradigms in Table~\ref{tab:unified_multitask}. Existing methods typically encounter a critical precision-sensitivity trade-off. For instance, tightly coupled shared-representation networks (e.g., Multi-task Transformers~\cite{zhou2021multi}) experience task interference, resulting in compromised diagnostic sensitivity (65.40\%). Conversely, conventional pipeline approaches (e.g., Bruno et al.~\cite{bruno2025dual}) isolate these tasks but remain vulnerable to error propagation from initial segmentation, leading to degraded boundary delineation (mDSC: 0.7690).

By substituting rigid feature-sharing with task-driven spatial prompts, our two-stage cascaded framework maintains high topological fidelity in segmentation (mDSC: 0.9075, mIoU: 0.8351). Furthermore, the integration of the dual-branch diagnostic head yields a balanced diagnostic profile with a Sensitivity of 97.62\% and an F1-score of 97.65\%. These quantitative gains indicate that explicitly fusing visually grounded deep representations with physical radiomics proxies effectively mitigates the task interference bottleneck inherent in previous joint-learning models.

\subsection{Ablation Study}
\subsubsection{Contribution Attribution of Architecture Components}

Table \ref{tab:unified_ablation} details the diagnostic contributions of the proposed dual-branch architecture and visual feature synergy modules. The ablation reveals that the Multi-pooling module is the primary contributor to diagnostic Sensitivity, exhibiting a 14.28\% absolute drop when removed. This confirms its critical role in extracting heterogeneous local features from the lesion core and proximal neighborhood. Meanwhile, the Adaptive ASPP primarily enhances global context modeling, evidenced by a 4.25\% improvement in AUC. 

Furthermore, integrating the explicit radiomics branch significantly mitigates the false-negative risk inherent in pure deep learning models (the SAM-only configuration), boosting Sensitivity from 88.89\% to 98.41\%. By transparently fusing these implicit visual representations with explicit physical priors within our two-stage cascaded pipeline, the framework avoids the optimization shortcuts commonly seen in end-to-end architectures, achieving an optimal diagnostic equilibrium.

\subsubsection{Ablation Study on the Multi-Box Fusion Strategy}\par
To validate the individual contributions and synergistic effects of our multi-box fusion strategy, we conducted a systematic ablation study excluding the PyRadiomics branch to isolate visual prompt variations (Table \ref{tab:multi_task_formatted}). We first compared prompt sources, revealing a critical trade-off between geometric precision and diagnostic utility. While the manually annotated GT-box achieves the upper bound in segmentation (mDSC 0.9270), its classification performance is surprisingly limited (AUC 86.06\%). In contrast, our weakly supervised CAM-box sacrifices marginal localization accuracy but substantially boosts classification metrics (AUC 91.34\%, F1 85.2\%). This indicates that strict geometric encapsulation of the lesion discards vital perilesional context, such as boundary transitions and local structural disturbances. The CAM-box naturally preserves this discriminative visual evidence, demonstrating that task-driven prompt generation is superior to pure anatomical localization for downstream diagnosis.

Building upon the CAM-box, we sequentially integrated our three fusion mechanisms to address spatial, semantic, and optimization robustness. Introducing V1 improves Accuracy to 84.75\% by mitigating the instability inherent to single-box thresholding. Adding V2 explicitly aligns candidate regions with textual descriptors, which strongly biases the model toward recalling malignant features and pushes Sensitivity to a peak of 92.06\%. Conversely, V3 acts as a robust denoising mechanism; it favors a more conservative decision boundary that maximizes Specificity (90.91\%) and AUC (92.47\%) by suppressing optimization noise from poor prompts. Ultimately, the joint integration of all components (V1+V2+V3) achieves the optimal diagnostic equilibrium (Accuracy 89.83\%, AUC 92.84\%, F1 90.32\%) without degrading segmentation quality (mDSC 0.9158). This confirms that our approach is not a naive prompt ensemble, but a cohesive, classification-aware framework that successfully balances spatial stability, semantic correctness, and training regularization.

\subsubsection{Ablation Study on Semantic Prompt Dependency and Robustness}

As detailed in Table \ref{tab:semantic_prompt}, the integration of correct fine-grained text prompts yields the highest diagnostic performance, confirming that semantic knowledge effectively acts as a spatial prior to guide precise lesion localization. More importantly, when subjected to random or deliberately mismatched prompts, the framework exhibits remarkable stability, maintaining an ACC above 93.22\% without catastrophic degradation.

\begin{table}[htbp]
  \centering
  \caption{Ablation study on semantic prompt dependency and robustness.}
  \label{tab:semantic_prompt}
  \begin{tabular}{l c c c}
    \toprule
    \textbf{Prompt Strategy} & \textbf{mDSC} & \textbf{mIoU} & \textbf{Acc (\%)} \\
    \midrule
    w/o Semantic Prompt & 0.9159 & 0.8506 & 90.68 \\
    Random Semantic Prompt & 0.9173 & 0.8506 & 93.22 \\
    Mismatched Prompt & 0.9113 & 0.8409 & 93.22\\
    \textbf{Correct Fine-grained Prompt (Ours)} & \textbf{0.9217} & \textbf{0.8579} & \textbf{94.07} \\
    \bottomrule
  \end{tabular}
\end{table}

These results suggest that the proposed framework is not overly dependent on the exact textual content of semantic prompts. Random or mismatched prompts do not cause catastrophic degradation, probably because textual semantics are used only for localization-oriented prompt construction and are not directly fed into the final diagnostic head. However, the performance gap between incorrect prompts and the correct fine-grained prompt is relatively small, indicating that the main diagnostic performance may also be strongly supported by the SAM visual branch and the radiomics branch, Therefore, the role of semantic prompting should be interpreted as improving localization stability rather than serving as the sole dominant factor for diagnosis.

\subsection{Interpretability and Feature Synergy Analysis}

\begin{figure}[htbp]
    \centering
    \includegraphics[width=\linewidth]{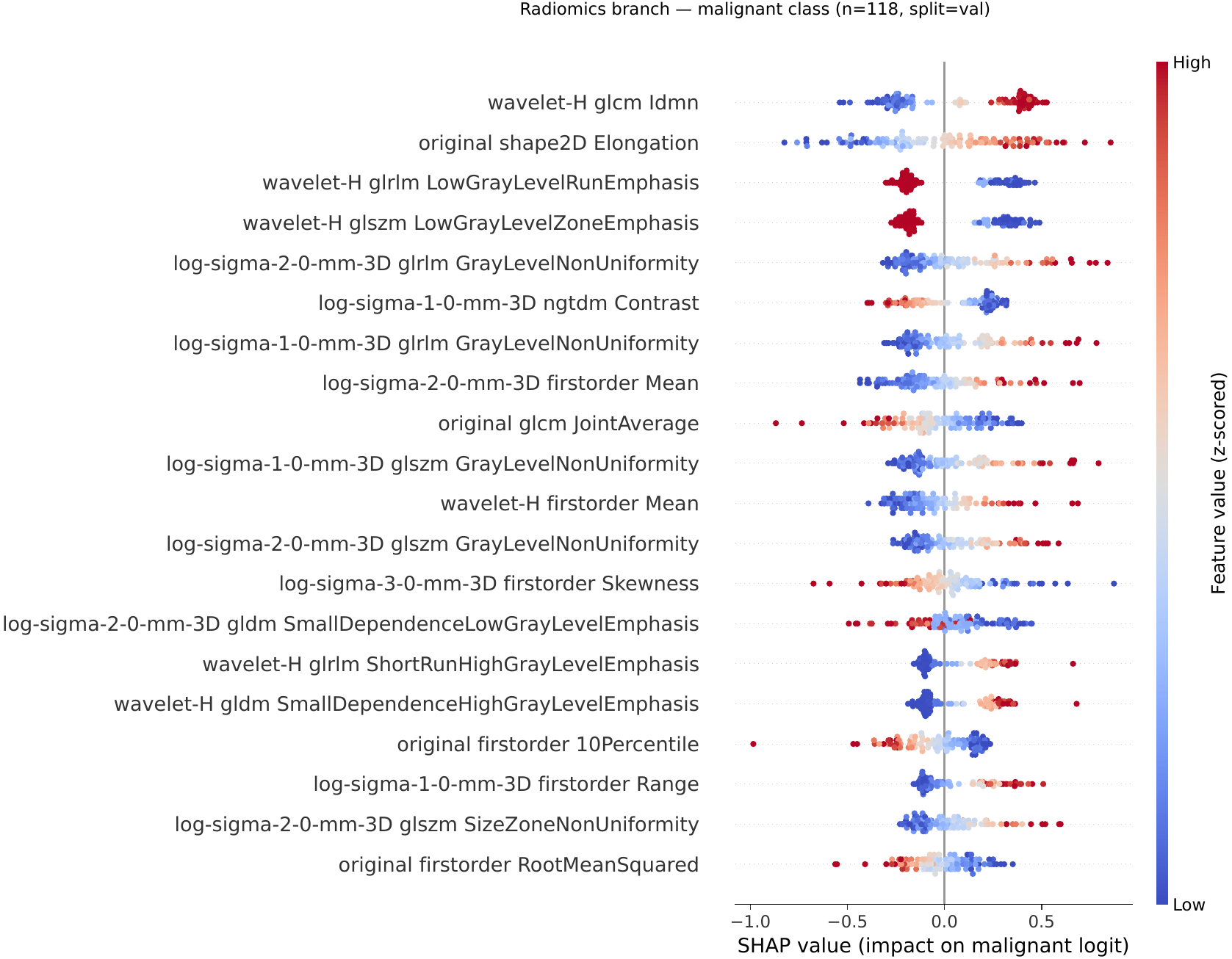}
    \caption{Global feature attribution via SHAP summary plot. The beeswarm plot reveals that texture-based radiomic features, notably \textit{wavelet-H glcm Idmn}, predominantly drive the malignant predictions.}
    \label{fig:shap_summary}
\end{figure}

To interpret the decision-making process of our fusion architecture and ensure clinical transparency, we employed SHapley Additive exPlanations (SHAP) for global feature attribution. As illustrated in the SHAP summary plot (Figure \ref{fig:shap_summary}), the model's predictions are predominantly driven by texture-based and filter-derived features that quantify intra-tumoral heterogeneity. Notably, \textit{wavelet-H glcm Idmn} emerges as the most decisive feature. Since the Inverse Difference Moment Normalized (IDMN) serves as an established metric for local homogeneity, its lower values—depicted by the blue cluster expanding towards the positive SHAP axis—significantly amplify the malignant logit. This observation aligns with the radiological consensus that malignant lesions typically exhibit highly heterogeneous and chaotic internal architectures. Furthermore, the strong prevalence of \textit{wavelet-H} and \textit{log-sigma} derived features underscores the model's reliance on multi-scale, high-frequency textural variations, capturing micro-environmental complexities that are often imperceptible to the naked eye.

\begin{figure}[htbp]
    \centering
    \includegraphics[width=\linewidth]{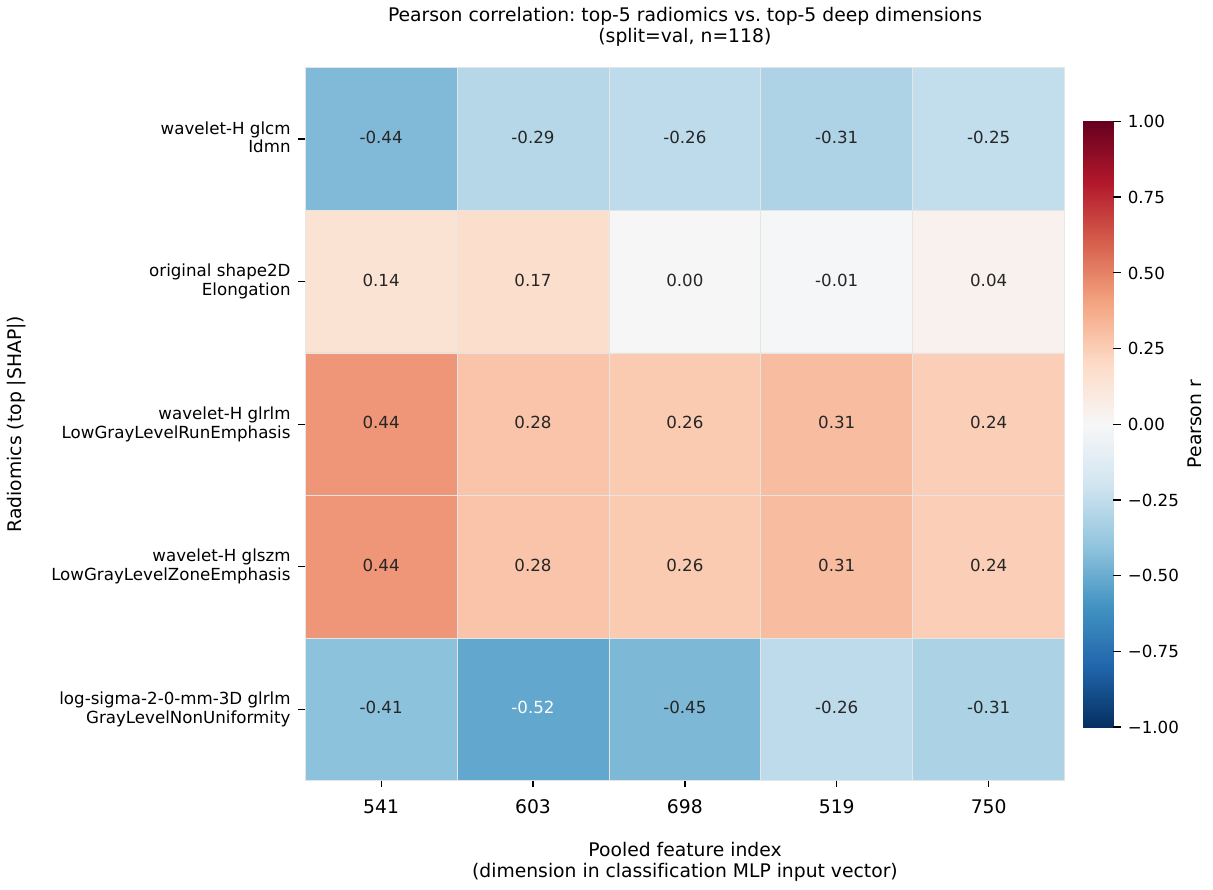}
    \caption{Cross-correlation heatmap between explicit radiomic priors and implicit deep representations. The analysis illustrates a moderate yet significant semantic alignment, highlighting the feature complementarity of the dual-branch framework.}
    \label{fig:corr_heatmap}
\end{figure}

Although SHAP clarifies the contributions of explicit omics features, the effectiveness of the model’s dual-branch fusion also depends on whether deep visual features capture complementary information; therefore, we further analyzed the cross-correlation between these two types of features. The cross-correlation heatmap (Figure \ref{fig:corr_heatmap}) reveals that the most discriminative dimensions of the pooled deep features exhibit moderate yet statistically significant correlations (with Pearson correlation coefficients peaking around 0.52) with top-ranking radiomic parameters, such as \textit{log-sigma-2-0-mm-3D glrlm GrayLevelNonUniformity}. This latent mapping provides a profound interpretability perspective: the deep neural network does not merely memorize abstract noise; rather, it implicitly encodes meaningful clinical phenotypes analogous to traditional heterogeneity metrics. Crucially, the moderate magnitude of these correlations demonstrates that the two feature spaces are fundamentally complementary rather than redundant. While the deep branch captures high-order, non-linear topological patterns, the radiomic pathway explicitly anchors the diagnostic reasoning with robust, physically interpretable constraints. Ultimately, this dual-stream synergy strongly supports the necessity of our proposed architecture in providing complementary information sources and bridging the gap between deep feature efficacy and clinical trust.

\section{Conclusion}
In this paper, we propose a two-stage multi-task framework to address annotation dependency and the lack of interpretability in medical image analysis. By organizing localization, prompt generation, segmentation, and diagnosis into a two-stage evidence-driven pipeline, our method utilizes a customized BLIP-2 architecture to automatically generate spatial priors, thereby significantly reducing manual prompt dependence. To ensure reliability, we introduce a MCRA strategy that mitigates imaging noise and semantic artifacts, substantially improving prompt robustness. Guided by these high-confidence prompts, a SAM-based decoder performs high-fidelity mask segmentation and binary classification, ensuring that the prediction supports a strictly visually grounded diagnosis. Finally, to enhance clinical interpretability, a dual-branch decision module synergizes deep visual features with physics-based radiomics proxies. Extensive experiments across five benchmarks demonstrate that our framework achieves state-of-the-art segmentation accuracy and an optimal diagnostic equilibrium.

\bibliographystyle{ieeetr}
\bibliography{sample}

\end{document}


\title{Supplementary Material for "Rad-VLSM:  A Cross-Modal Framework with Semantics-Assisted Prompting for Medical Segmentation and Diagnosis"}

\author{Fengyi Zhang\textsuperscript{*}, \IEEEmembership{Student Member, IEEE}, 
        Xujie Zeng\textsuperscript{*}, \IEEEmembership{Student Member, IEEE}, Mohan Liu\textsuperscript{*}, Zengyi Wang, \IEEEmembership{Student Member, IEEE} and Yalong Jiang, \IEEEmembership{Member, IEEE}

\thanks{Manuscript received April xx, 2026. (Corresponding author: Yalong Jiang.)}%
\thanks{Fengyi Zhang, Xujie Zeng, and Mohan Liu contributed equally to this work and are co-first authors.}%
\thanks{Fengyi Zhang, Xujie Zeng, Zengyi Wang, and Yalong Jiang are with the Department of Electronic Science and Technology, Hainan University, Haikou 570228, China (e-mail: zhangfengyi@hainanu.edu.cn; zengxujie@hainanu.edu.cn; 3214374014@qq.com; AllenYLJiang@outlook.com).}%
\thanks{Mohan Liu is with the Peking Union Medical College Hospital, Beijing 100730, China (e-mail: liumohan98@163.com).}}

\maketitle

\beginsupplement

\section{Introduction}
This document provides supplementary material for the main manuscript, ”Rad-VLSM: A Cross-Modal Framework with Semantics-Assisted Prompting for Medical Segmentation and Diagnosis.“ Due to space constraints in the main text, this document includes comprehensive methodological details and additional empirical analyses. Specifically, Section S2 elaborates on the rigorous three-stage cascaded feature selection mechanism—encompassing variance filtering and Z-score standardization, Minimum Redundancy Maximum Relevance (mRMR) pre-filtering, and LASSO-based absolute selection with sparsity constraints—designed to mitigate the curse of dimensionality in high-dimensional radiomics pools. Furthermore, Section S3 presents an in-depth ablation study that quantitatively validates the synergistic contributions of these individual selection components to the overall diagnostic performance.

\section{Additional Method Studies}
\subsection{Feature Selection and Dimensionality Reduction}
To mitigate the curse of dimensionality and minimize multicollinearity within the over 500-dimensional feature pool, we implement a rigorous three-stage cascaded feature selection mechanism to extract the most discriminative and non-redundant core radiomics signatures. This mechanism consists of three sequential steps: 1) Variance filtering and Z-score standardization for statistical stability; 2) Minimum Redundancy Maximum Relevance (mRMR) for redundancy pre-filtering; and 3) LASSO regression with strict sparsity constraints for absolute selection.

\subsubsection{Variance Filtering and Z-score Standardization}
Before formal feature selection, the raw radiomics features undergo a two-step preprocessing phase to ensure statistical stability. First, we apply a variance-based filter to eliminate near-constant features that lack discriminative power. A feature $x_f$ is retained only if its variance $\sigma^2(x_f)$ exceeds a predefined threshold $\tau$ calculated on the training set

Subsequently, we perform Z-score standardization to transform the features into a uniform scale, facilitating balanced optimization during downstream selection. The standardized feature $z_i$ is computed as:
\begin{equation}
z_i = \frac{x_i - \mu_{\text{train}}}{\sigma_{\text{train}}}
\end{equation}
where $\mu_{\text{train}}$ and $\sigma_{\text{train}}$ represent the mean and standard deviation of the corresponding feature in the training split, respectively.

\subsubsection{Redundancy Pre-filtering via mRMR}
Initially, the high-dimensional feature pool is filtered using the Minimum Redundancy Maximum Relevance (mRMR) algorithm. This mechanism maximizes the mutual information between features and the ground-truth labels while simultaneously suppressing redundancy by penalizing high Pearson correlation coefficients among selected features. Given a candidate feature set $\Omega$ and a selected subset $\mathcal{S}$, the evaluation score $\Phi(x_j)$ for an unselected feature $x_j \notin \mathcal{S}$ is defined as:
\begin{equation}
\Phi(x_j) = I(x_j; y) - \frac{1}{|\mathcal{S}|} \sum_{x_k \in \mathcal{S}} \left| \rho(x_j, x_k) \right|
\end{equation}
where $I(x_j; y)$ denotes the mutual information between feature $x_j$ and label $y$, and $\rho(x_j, x_k)$ represents the correlation coefficient between $x_j$ and the previously selected feature $x_k$. Utilizing a greedy strategy, the system iteratively retains the top $K$ robust features.

\subsubsection{Absolute Selection via LASSO with Sparsity Constraints}
Building upon the mRMR reduction, we further introduce a Logistic Regression model with $L_1$ regularization (LASSO) for fine-grained feature refinement. By imposing strict sparsity constraints, LASSO forces the weights of features with negligible contributions or high collinearity toward zero. The objective function is formalized as:
\begin{equation}
\min_{w} \left[ - \frac{1}{N} \sum_{i=1}^{N} \log P(y_i | x_i; w) + \lambda \| w \|_1 \right]
\end{equation}
where $N$ is the number of training samples, $w$ is the feature weight vector, and $\lambda$ is the optimal regularization coefficient determined through adaptive search via cross-validation. Ultimately, only core radiomics features with non-zero coefficients are retained to construct the final radiomics descriptor.

Specifically, on our custom breast ultrasound dataset, this rigorous three-stage mechanism successfully reduced the initial pool of over 500 extracted features down to exactly 63 highly discriminative and non-redundant radiomics proxies, which are subsequently utilized in the dual-branch decision module

\begin{table*}[t] 
    \centering
    \caption{Ablation study of the proposed cascaded radiomics feature selection mechanism on the clinical dataset. The results demonstrate the synergistic effect of normalization, redundancy filtering, and sparsity constraints. Best results are highlighted in bold.}
    \label{tab:ablation_selection_fullwidth}
    \renewcommand{\arraystretch}{1.3} 
    
    \begin{tabularx}{\textwidth}{@{\extracolsep{\fill}} ccc S[table-format=2.2] S[table-format=2.2] S[table-format=2.2] S[table-format=2.2] S[table-format=2.2] @{}}
        \toprule
        \multicolumn{3}{c}{\textbf{Methodological Components}} & \multicolumn{5}{c}{\textbf{Evaluation Metrics (\%)}} \\ 
        \cmidrule(lr){1-3} \cmidrule(lr){4-8}
        \textbf{Z-score} & \textbf{mRMR} & \textbf{LASSO} & {\textbf{ACC} $\uparrow$} & {\textbf{AUC} $\uparrow$} & {\textbf{Sens.} $\uparrow$} & {\textbf{Spec.} $\uparrow$} & {\textbf{F1} $\uparrow$} \\ 
        \midrule
        $\times$     & $\times$     & $\times$     & 87.29 & 92.35 & 90.48 & 83.64 & 88.37 \\
        $\checkmark$ & $\times$     & $\times$     & 90.68 & 96.42 & 92.06 & 89.09 & 91.34 \\
        $\checkmark$ & $\checkmark$ & $\times$     & 92.37 & 96.97 & 96.83 & 87.27 & 93.13 \\
        $\checkmark$ & $\times$     & $\checkmark$ & 91.53 & 96.94 & 95.24 & 87.27 & 92.31 \\
        \midrule
        \rowcolor{gray!10} 
        $\checkmark$ & $\checkmark$ & $\checkmark$ & \textbf{94.07} & \textbf{97.55} & \textbf{98.41} & \textbf{89.09} & \textbf{94.66} \\
        \bottomrule
    \end{tabularx}
\end{table*}

\section{Additional Ablation Studies}
\subsection{Radiomics Feature Selection Mechanism}
To effectively integrate physical radiomics proxies without triggering the curse of dimensionality, we employ a three-stage selection mechanism (Table \ref{tab:ablation_selection_fullwidth}). Utilizing raw, unscaled radiomics features yields suboptimal performance (AUC 92.35\%) due to scale disparities and uninformative noise. While basic standardization (Z-score) partially rectifies this, significant inter-feature redundancy remains.

The sequential integration of mRMR and LASSO progressively distills the feature pool. Specifically, mRMR captures diverse malignant patterns by maximizing mutual information, which noticeably elevates Sensitivity to 96.83\%. Subsequently, LASSO enforces strict sparsity to prune weak predictors and collinear noise. As shown in the complete pipeline configuration, this systematic cascade successfully reduces over 500 raw features into a highly discriminative set of 63 physical priors. This refined descriptor seamlessly complements the deep visual logits, establishing the optimal diagnostic equilibrium across all metrics (Accuracy 94.07\%, AUC 97.55\%).

\subsection{Impact of Spatial Priors on Diagnostic Reasoning}
To dissect the contribution of lesion spatial evidence to the downstream diagnostic head, we conducted a granular ablation on the multi-dimensional pooling strategy (Table \ref{tab:spatial_prior_ablation}). The baseline without any spatial prior (S1) yields the lowest overall metrics, confirming that global visual features alone are insufficient to overcome pervasive ultrasound speckle noise. 
\begin{table}[htbp]
    \centering
    \caption{Ablation study on the spatial prior contributions to the diagnostic branch. The results demonstrate that combining the predicted lesion mask with bounding box context (S3) yields the optimal diagnostic profile, surpassing even the ground-truth mask baseline (S4).}
    \label{tab:spatial_prior_ablation}
    \renewcommand{\arraystretch}{1.2}
    \setlength{\tabcolsep}{4pt} 
    \begin{tabularx}{\columnwidth}{@{\extracolsep{\fill}} l ccccc @{}}
        \toprule
        \textbf{Configuration} & \textbf{ACC} $\uparrow$ & \textbf{AUC} $\uparrow$ & \textbf{Sens.} $\uparrow$ & \textbf{Spec.} $\uparrow$ & \textbf{F1} $\uparrow$ \\ 
        \midrule
        S1: No Prior               & 83.90 & 91.75 & 82.54 & 85.45 & 84.55 \\
        S2: Box Prior Only         & 86.44 & 92.35 & 87.30 & 85.45 & 87.30 \\
        S5: Predicted Mask Only    & 91.53 & 95.21 & \textbf{98.41} & 83.64 & 92.54 \\
        S4: GT Mask Prior          & 92.37 & 96.65 & 95.24 & \textbf{89.09} & 93.02 \\
        \midrule
        \textbf{S3: Ours (Rad-VLSM)} & \textbf{94.07} & \textbf{97.55} & \textbf{98.41} & \textbf{89.09} & \textbf{94.66} \\
        \bottomrule
    \end{tabularx}
\end{table}
Deconstructing the spatial priors reveals a clear functional complementarity. Relying exclusively on the predicted lesion mask (S5) maximizes Sensitivity (98.41\%) but degrades Specificity (83.64\%), indicating that isolating the lesion core neglects vital perilesional information, leading to increased false positives. By integrating the bounding box prior alongside the mask (S3), our proposed configuration successfully restores Specificity to 89.09\% without compromising Sensitivity. This quantitative shift demonstrates that the bounding box effectively captures critical perilesional infiltration context (e.g., spiculated margins and acoustic shadowing) required to accurately distinguish complex benign mimics.

Crucially, the proposed method relying on the predicted mask (S3) outperforms the theoretical baseline utilizing the manual Ground-Truth mask (S4) across multiple metrics, notably improving F1-score from 93.02\% to 94.66\%. This phenomenon suggests that rather than acting as a rigid geometric constraint, the network-predicted mask functions as a data-driven soft-attention mechanism. It implicitly accommodates the subtle transitional gradients at the tumor boundary, aligning more naturally with the shared deep feature representations than the rigid anatomical boundaries delineated by human annotators. Consequently, this design not only demonstrates robustness against upstream segmentation noise but actively leverages it for richer feature extraction.